\documentclass{article}

\PassOptionsToPackage{numbers,sort&compress}{natbib}
\usepackage[preprint]{neurips_2026}

\usepackage[utf8]{inputenc}
\usepackage[T1]{fontenc}
\usepackage{hyperref}
\usepackage{url}
\usepackage{booktabs}
\usepackage{graphicx}
\usepackage{amsmath,amssymb}
\usepackage{amsfonts}
\usepackage{nicefrac}
\usepackage{microtype}
\usepackage{xcolor}

\title{When Attention Does Not Explain the Peak:\\
Temporal Reference vs. Forecast Output in Attention-Based Time-Series Forecasting}

\author{
Yuji Akamatsu \qquad Takao Yamanaka \\
Sophia University \\
Tokyo, Japan
}

\newcommand{\psio}{\Psi_{\mathrm{out}}}

\newcommand{\htrue}{h_{\mathrm{true}}}
\newcommand{\hpred}{h_{\mathrm{pred}}}
\newcommand{\hpsi}{h_{\Psi}}

\begin{document}
\maketitle

\begin{abstract}
Attention maps are often interpreted as evidence of what a forecasting model
uses when making predictions. In our load-forecasting model, a CLS
representation of historical demand queries 24 future exogenous horizon tokens
through cross-attention, inviting a temporal interpretation in which highly
attended horizons may appear to explain forecast peak timing. We test this
interpretation using a horizon-level attention descriptor, $\psio$. Across 31
day-aligned windows of the Panama load dataset, the forecast achieves a median
peak-time error of 0 h and a 51.6\% exact-match rate, whereas the argmax of
$\psio$ has a median error of 5 h and 0\% exact match. The forecast peak is
closer to the observed peak in 27 of 31 windows. This dissociation is not merely
an argmax artifact: within $\pm1$ h, attention reaches only $1.16\times$,
$1.11\times$, and $1.14\times$ the uniform baseline around observed,
predicted, and weekly-naive peaks, respectively, indicating weak and
non-selective concentration. Yet the attention profile is structured, with
cross-window consistency of 0.83. Replacing 12 future weather features with
their training-set means makes the profile nearly uniform, showing sensitivity
to future weather variation rather than fixed horizon position alone. The
dissociation is also reproduced across three random-seed runs. These results
show that structured, input-sensitive, and reproducible horizon-level
cross-attention need not provide a valid peak-selective explanation of forecast
behavior. The observed behavior is instead consistent with an internal
horizon-reference role for integrating future exogenous information, although
this functional role is not causally established.
\end{abstract}

\section{Introduction}

Short-term electric load forecasting is important for power-system planning
and operations~\cite{hong2016load}. Within this setting, peak-demand timing is
operationally relevant because a forecast can achieve acceptable aggregate
error while still misplacing a salient peak in time. When attention
maps are used to interpret such forecasts, it becomes important to distinguish
whether the attention pattern merely has internal structure or actually
supports the specific output-level interpretation assigned to it. In classical statistical time-series models, the relationship between model components and forecast outputs is often specified explicitly through equations and model assumptions. In deep forecasting models, by contrast, observable internal quantities such as attention weights do not necessarily provide a direct explanation of specific forecast outputs.

Point-wise error metrics such as RMSE and MAE summarize forecasting accuracy in the value space but do not isolate temporal misalignment of salient events. For peak-demand forecasting, we therefore use peak-time error ($\Delta t_{\rm peak}$) as a complementary event-level diagnostic that measures the difference between observed and predicted peak horizons in physical time units. This diagnostic is computed only from forecast outputs and observations; the attention descriptor $\psio$ is treated separately and is never used to define $\Delta t_{\rm peak}$ or to select models.

Our question arises from the specific cross-attention structure of the forecasting model. A 168-step history window is rendered as an image and encoded by a Vision Transformer. The resulting CLS representation, which summarizes the historical demand state, then queries 24 future exogenous horizon tokens through a cross-attention module. We define $\psio$ as the head-averaged cross-attention weight over these future horizons. Thus, $\psio$ indicates how the historical state references the future exogenous horizon tokens during forecasting; it is not self-attention over image patches and not a per-covariate importance score. Because the historical CLS state directly attends to future horizon tokens,
the resulting attention profile admits a tempting temporal interpretation:
horizons receiving larger attention may appear to contribute more strongly
to forecast formation. Such an interpretation is plausible because attention
is assigned directly over future forecast horizons, even though these tokens
represent exogenous variables rather than demand outputs. We test one observable implication of a peak-selective interpretation:
whether $\psio$ aligns with forecast peak timing. Alignment with the observed
peak provides an additional reference for whether the attended horizon
coincides with the salient event that the forecast is intended to reproduce.
We therefore ask a deliberately narrow question:
\emph{can horizon-level cross-attention be interpreted as an explanation
of forecast peak timing?}

We find a clear dissociation between forecast-output timing and this internal horizon reference. The forecast peak is closely aligned with the observed peak, whereas the argmax of $\psio$ is not. A peak-neighborhood mass analysis shows that this result is not explained by instability of the attention argmax: $\psio$ is only weakly concentrated around peak horizons and is similarly concentrated around observed, predicted, and weekly-naive peaks. The more important finding, however, is that this mismatch does not arise because the attention profile is arbitrary. $\psio$ is highly consistent across test windows, revealing a stable temporal structure rather than unstructured noise. Moreover, replacing future weather features with their training-set means causes this structure to become nearly uniform. Thus, the attention profile is sensitive to variation in future weather features even though it remains weakly aligned with demand-peak timing. Together with the architecture, in which the historical CLS representation queries future exogenous horizon tokens, this behavior is consistent with a role as an internal horizon reference for integrating future exogenous information rather than as a selector of the demand-peak horizon. We treat this as an interpretation of the observed sensitivity. The central forecast--attention dissociation persists across three independently trained random seeds.

Our contribution is therefore not a new forecasting model or attention
mechanism, but an evaluation of an interpretive assumption induced by the
model's cross-attention structure.
In contrast to much prior work that evaluates attention primarily in terms of
feature or token relevance, we directly test whether horizon-level
cross-attention over future exogenous inputs aligns with a concrete output
event: forecast peak timing.
The results distinguish four properties that can otherwise
be conflated: attention structure, input sensitivity,
diagnostic utility, and output-level explanatory validity.
More generally, attention patterns should be validated
against the specific output behavior they are claimed
to explain rather than treated as explanatory from
their structure alone.

\section{Related Work}

\paragraph{Temporal evaluation in forecasting.}
Point-wise metrics such as RMSE and MAE summarize value-space
forecasting accuracy but do not directly characterize the temporal
alignment of salient events. Le Guen and Thome~\cite{leguen2019dilate}
make a related distinction in multi-step forecasting, separating
shape error from temporal distortion in both their evaluation and
their proposed training loss. While their work aims to improve forecast
trajectories, we use peak timing as an event-level diagnostic of the
temporal relationship between forecast outputs and attention.

\paragraph{Seasonal-naive controls.}
Electricity demand exhibits strong periodic structure, so a weekly
seasonal-naive predictor~\cite{hyndman2021fpp} provides a natural
periodicity control for peak timing. Comparing attention alignment
against this control helps assess whether alignment with the model's
forecast peaks is distinguishable from alignment with peaks implied
by recurring weekly demand patterns.

\paragraph{Attention as explanation.}
Whether attention weights constitute explanations of model predictions
remains contested. Jain and Wallace~\cite{jain2019attention}
find that attention weights often correlate poorly with
gradient-based importance measures and that substantially
different attention distributions can yield similar predictions.
Serrano and Smith~\cite{serrano2019attention} show that
attention weights are imperfect indicators of the importance
of input representations to model decisions.
Wiegreffe and Pinter~\cite{wiegreffe2019attention} argue that
attention may nevertheless provide plausible explanations,
depending on how explanations are defined and evaluated.
Bibal et al.~\cite{bibal2022attention} review this debate
and emphasize that the explanatory role of attention
depends on the intended notion of explanation.
Much of the foundational debate focused on attention
in NLP tasks, particularly recurrent architectures,
rather than horizon-level cross-attention in time-series forecasting.

\paragraph{Attention in time-series forecasting.}
Attention serves different roles in forecasting architectures.
The Temporal Fusion Transformer~\cite{lim2021tft}
uses interpretable temporal self-attention to analyze
learned temporal dependencies across past and known-future inputs.
TimeXer~\cite{wang2024timexer} integrates exogenous information
through cross-attention from an endogenous global token to
variate-level exogenous tokens. Unlike our setting, its
exogenous tokens represent historical variables rather
than individual future horizons.
These architectures illustrate the use of attention
for temporal interpretation and exogenous integration,
respectively.
In the broader time-series interpretability literature,
Ismail et al.~\cite{ismail2020benchmarking} show that
saliency methods performing well on images can fail to
identify informative time steps and features in multivariate
time series, motivating quantitative evaluation of
explanation quality.
Our study does not propose a competing mechanism; it
complements these works by examining whether the temporal
location of horizon-level cross-attention aligns with a
specific forecast output event: the predicted peak.

\paragraph{Research gap and evaluation scope.}
Prior work motivates a distinction between the structure of an
attention pattern and its validity as an explanation of a specific
output. In our setting, a single historical CLS query attends to
24 future exogenous horizon tokens, so the most-attended horizon
may be interpreted as the future time step most strongly
referenced during forecast formation. We test whether this
interpretation is supported for forecast peak timing.
We characterize the attention profile through cross-window
consistency, input sensitivity, and random-seed robustness,
and evaluate its selectivity for peak timing through
peak-time alignment and local attention mass, including
a weekly seasonal-naive periodicity control.
As a secondary analysis, we examine whether
forecast--attention disagreement provides diagnostic
information about peak-timing failure.

\section{Method}

\subsection{Model and horizon reference descriptor}

The forecasting model renders a 168-step history window as a Gramian Angular
Field image~\cite{wang2015gaf}, encodes it with a ViT-B/16
backbone~\cite{dosovitskiy2021vit}, and fuses 24 future exogenous
horizon tokens through a cross-attention block. The image-derived CLS state is
used as the query, while the future horizon tokens serve as keys and values. A
regression head produces a 24-step forecast.

For each forecast origin, the 24 future-horizon tokens correspond to the
24 hours of the forecast horizon. Each horizon token contains 15 exogenous
variables for the corresponding future hour: four weather variables
(T2M, QV2M, TQL, and W2M) at each of three locations (toc, san, and dav),
together with three calendar-related variables (Holiday-ID, holiday, and
school). No future demand value is included in these tokens. The weather values are historical reanalysis records from the dataset,
so the analysis assumes perfect foresight of future weather; calendar
variables are known in advance.

Let $A^{(m)}(h)$ denote the attention weight from the CLS query to future horizon
token $h$ in head $m$, where $h\in\{1,\ldots,24\}$. We define the horizon
reference descriptor as
\begin{equation}
  \psio(h) = \frac{1}{M}\sum_{m=1}^{M} A^{(m)}(h),
  \qquad h=1,\ldots,24,
\end{equation}
where $M$ is the number of attention heads. Because the weights are normalized
by the attention softmax, $\psio$ sums to one over the 24 horizons. It is used
only as an analysis descriptor and does not alter training or inference.

\subsection{Peak timing}

For each day-aligned 24-hour forecast window, we define
\begin{align}
  \htrue &= \arg\max_h y(h), \\
  \hpred &= \arg\max_h \hat{y}(h), \\
  \hpsi  &= \arg\max_h \psio(h),
\end{align}
with ties resolved to the earliest horizon. Peak-time errors are
\begin{equation}
  e_{\mathrm{pred}} = |\hpred-\htrue|,
  \qquad
  e_{\Psi} = |\hpsi-\htrue|.
\end{equation}
We additionally report exact-match rates. As a periodicity control, we use the
weekly seasonal-naive peak obtained with lag 168 h.

\subsection{Peak-neighborhood mass}

Because the argmax of a diffuse attention profile can be unstable, we also
measure how much attention mass lies near a peak. For center horizon $c$ and
radius $k$,
\begin{equation}
  \mathrm{Mass}_k(c)
  = \sum_{h=\max(1,c-k)}^{\min(24,c+k)} \psio(h).
\end{equation}
We normalize this quantity against the corresponding uniform mass:
\begin{equation}
  \mathrm{Lift}_k(c)
  =
  \frac{\mathrm{Mass}_k(c)}
  {(\min(24,c+k)-\max(1,c-k)+1)/24}.
\end{equation}
The denominator is the attention mass expected within the same neighborhood
under a uniform distribution over the 24 horizons. Thus, a lift of 1
corresponds to uniform attention, whereas values greater than 1 indicate
increasing concentration around the target horizon.

\subsection{Window consistency}

To distinguish a stable attention structure from window-specific peak tracking,
we compute the mean correlation between each window's attention profile and the
window-averaged profile:
\begin{equation}
  \mathrm{Consist}
  =
  \frac{1}{N}\sum_{i=1}^{N}
  \rho\!\left(\Psi_{\mathrm{out},i},\bar{\Psi}_{\mathrm{out}}\right),
  \qquad
  \bar{\Psi}_{\mathrm{out}}
  =
  \frac{1}{N}\sum_{i=1}^{N}\Psi_{\mathrm{out},i}.
\end{equation}
Here, $\rho$ denotes the Pearson correlation coefficient.
Higher consistency indicates that the attention profile is similar across test
windows rather than adapting strongly to each window's peak location.

\subsection{Weather-input sensitivity probe}

To examine whether the horizon-level attention structure reflects only a
fixed positional pattern or is sensitive to future exogenous content, we
perform a test-time sensitivity probe. Keeping the trained model fixed, we
replace the 12 weather features (T2M, QV2M, TQL, and W2M at the three
locations) in the future horizon tokens with their training-set means and
re-run inference while leaving the calendar-related inputs unchanged. We then
recompute the horizon-level attention profiles and peak-neighborhood mass.

This experiment is intended only to measure sensitivity to variation in the
future weather features. It is not a causal intervention and is not used for
model selection or training.

\subsection{Data and protocol}

We evaluate on the Panama electricity-demand dataset~\cite{aguilar2021panama_dataset}. January 2019 is held out
as the test period, yielding 31 day-aligned 24-hour forecast windows in the
primary analysis. Normalization statistics are estimated from training data
only. All peak-timing and attention diagnostics are computed on these test
windows.

For the random-seed robustness analysis, we use the 30 day-aligned origins that
are available identically in the saved test outputs of all three independently
trained runs. The timestamps are checked programmatically to ensure that the
same evaluation origins are compared across seeds.

\section{Results}

\subsection{Forecast and attention peaks are dissociated}

\begin{figure}[t]
  \centering
  \includegraphics[width=0.76\linewidth]{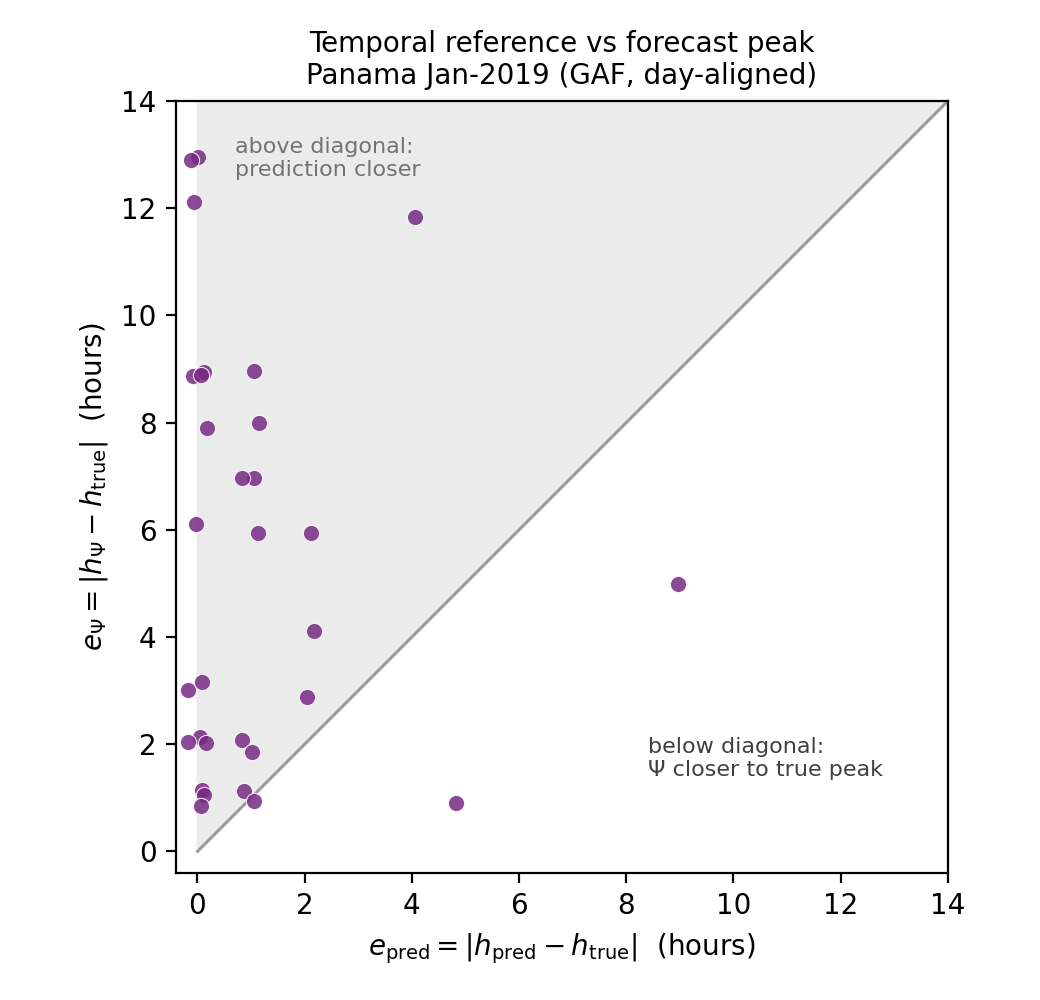}
  \caption{Dissociation between forecast peak error and attention-derived peak
  error over 31 day-aligned windows. Points above the diagonal indicate windows
  in which the forecast peak is closer to the observed peak than the
  $\psio$ argmax. This occurs in 27 of 31 windows.}
  \label{fig:dissociation}
\end{figure}

Over the 31 day-aligned windows, the forecast attains a median peak-time error
of 0 h and a 51.6\% exact-match rate, matching the weekly seasonal-naive
baseline on these two coarse statistics. The baseline serves here primarily
as a periodicity control rather than as a forecasting benchmark: the question
is whether attention aligns with recurring weekly structure, for which it
shows only weak local concentration (lift $1.14\times$;
Section~\ref{sec:argmax_artifact}).
In contrast, the argmax of $\psio$
has a median error of 5 h and never exactly matches the observed peak.
Window by window, the forecast peak is closer to the observed peak than the
$\psio$ argmax in 27 of 31 windows; $\psio$ is closer in 2 windows, with 2 ties.
This difference is also supported by a Wilcoxon signed-rank test
($p<0.001$).
These results do not support interpreting the attention argmax as a direct
marker of where the model places its forecast peak
(Figure~\ref{fig:dissociation}; Table~\ref{tab:main_results}). Representative demand windows illustrating the forecast--attention dissociation are provided in Appendix~\ref{app:representative_waveforms}.

\subsection{The discrepancy is not an argmax artifact}
\label{sec:argmax_artifact}

Because the argmax of a diffuse attention profile can be unstable, we next measure attention mass within $\pm 1$ h of each peak. Relative to a uniform profile, $\psio$ reaches $1.16\times$ around the observed peak, $1.11\times$ around the predicted peak, and $1.14\times$ around the weekly-naive peak. These values indicate only weak concentration and, more importantly, no clear selectivity toward either the observed or predicted peak. The mismatch between attention and peak timing is therefore not explained solely by instability of the attention argmax.

\begin{table}[t]
\centering
\caption{Peak-time comparison on the Panama test period (31 day-aligned windows).}
\label{tab:main_results}
\begin{tabular}{lcc}
\toprule
Signal & Median peak error (h) & Exact match \\
\midrule
Forecast (ViT, GAF)    & 0 & 51.6\% \\
Weekly naive (lag 168) & 0 & 51.6\% \\
$\psio$ argmax         & 5 & 0.0\% \\
\bottomrule
\end{tabular}
\end{table}

\subsection{The attention profile is structured but not peak-selective}

Despite its weak alignment with individual peaks, $\psio$ is not unstructured noise. Individual attention profiles are highly consistent with the window-averaged profile, with a window-consistency of 0.83. The average profile is smooth and broadly diurnal rather than sharply concentrated around window-specific peak locations.

Taken together, these results show that $\psio$ contains a stable temporal structure, but that this structure does not selectively track the timing of either the forecast or observed peak. The absence of peak selectivity therefore should not be conflated with an absence of internal structure.

\subsection{The attention structure is sensitive to future exogenous variation}

\begin{figure}[t]
\centering
\includegraphics[width=0.92\linewidth]{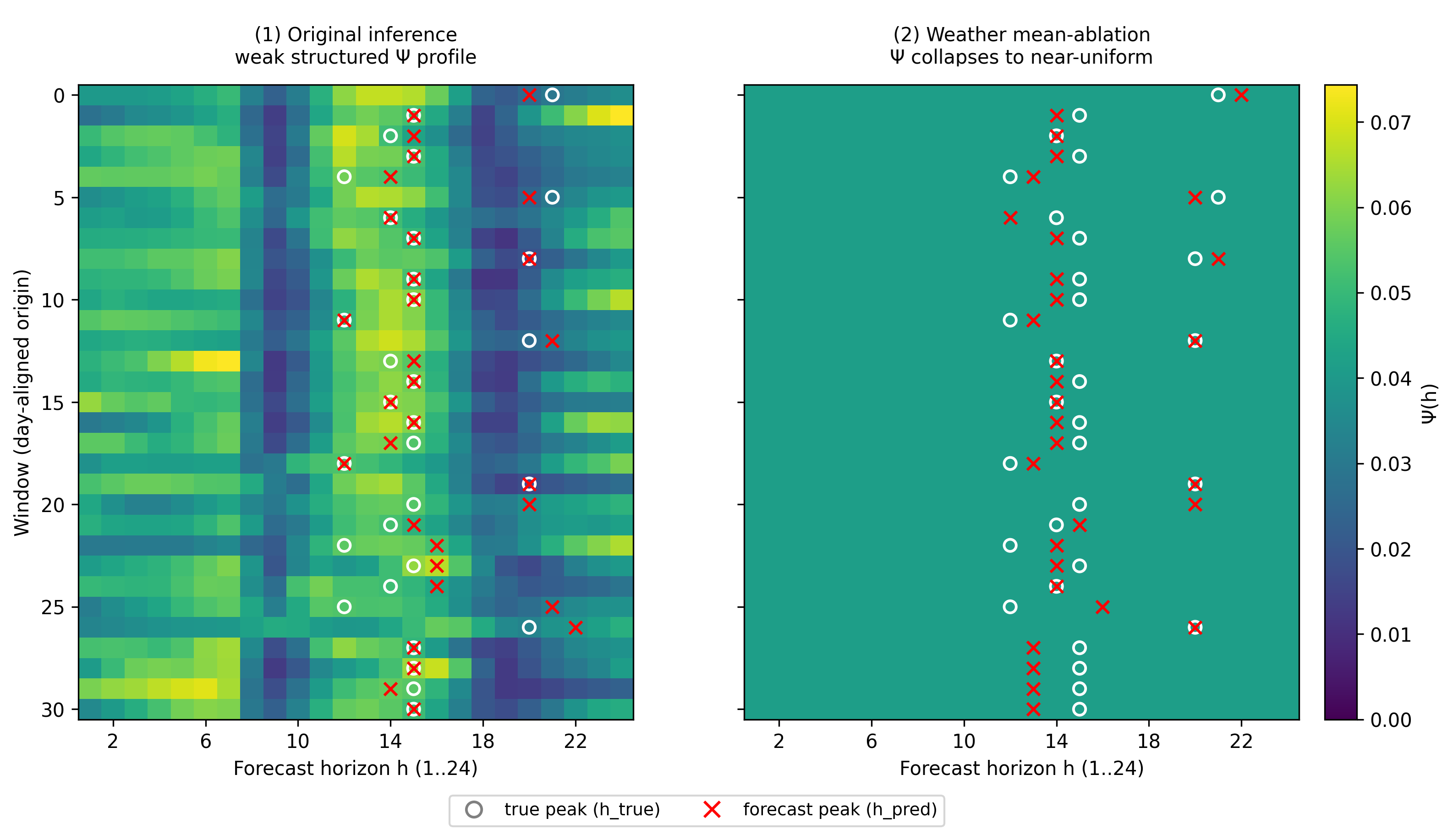}
\caption{Horizon-level attention profiles under the original inference
setting (left) and the weather-mean sensitivity condition (right).
Rows correspond to test windows and columns to the 24 future horizons.
Under the original inputs, $\psio$ exhibits a structured and broadly
consistent temporal profile that does not selectively coincide with
the observed or forecast peaks. Replacing future weather features
with their training-set means makes the profile nearly uniform,
showing sensitivity to changes in future weather inputs rather than
a fixed horizon-position pattern alone.}
\label{fig:horizon_attention}
\end{figure}

To probe what gives rise to the stable horizon-level attention structure, we replaced the future weather features with their training-set means at test time. Under this condition, the structured attention profile became nearly uniform, and the $\pm 1$ h lift around the observed peak decreased from $1.16$ to $1.00$ (Figure~\ref{fig:horizon_attention}).

This result indicates that the horizon-level attention structure is sensitive
to variation in the future weather features and cannot be explained by a fixed
positional template alone. Together with the cross-attention architecture, in which the historical CLS state queries future exogenous horizon tokens, the observed pattern is consistent with $\psio$ serving as an internal horizon reference for integrating future exogenous information.

\subsection{The dissociation is consistent across random seeds}

We next test whether the output--attention dissociation depends on a particular
random initialization. We repeat the analysis for three independently trained runs initialized with different random seeds using the same 30 day-aligned
evaluation origins. Because this robustness analysis is restricted to the
30 origins common to all three runs, the seed-0 summary differs slightly from
the 31-window primary analysis above.

The forecast median peak-time error is 0 h for all three runs, whereas the
median error of the $\psio$ argmax is 4.5, 6.0, and 1.0 h for seeds 0, 1,
and 2, respectively. The forecast peak is closer to the observed peak than
the attention-derived peak in 26/30, 25/30, and 22/30 windows. In contrast,
$\psio$ is closer in only 2 windows for each seed. The paired differences are
supported by Wilcoxon signed-rank tests for all three seeds
($p=0.0001$, $p<0.0001$, and $p=0.0004$, respectively). The corresponding results are summarized in Table~\ref{tab:seed_robustness}.

\begin{table}[t]
\centering
\caption{Random-seed robustness on the 30 day-aligned origins common to all three runs. $\psio$/F denotes the numbers of windows in which the attention-derived peak and the forecast peak, respectively, are closer to the observed peak.}
\label{tab:seed_robustness}
\begin{tabular}{lcccccc}
\toprule
Seed &
$e_{\mathrm{pred}}$ &
$e_{\Psi}$ &
Forecast exact &
$\psio$ exact &
$\psio$/F &
Consistency \\
\midrule
0 & 0.0 & 4.5 & 53.3\% & 0.0\%  & 2/26 & 0.82 \\
1 & 0.0 & 6.0 & 53.3\% & 10.0\% & 2/25 & 0.79 \\
2 & 0.0 & 1.0 & 66.7\% & 16.7\% & 2/22 & 0.89 \\
\bottomrule
\end{tabular}
\end{table}

The attention profiles remain structurally consistent across windows, with consistency values between 0.79 and 0.89. The observed-peak neighborhood lift, however, varies more across seeds, from 1.12 to 1.56. We therefore do not claim that the degree of local peak concentration is invariant to initialization. The more direct result is robust: for every seed, the forecast peak is substantially more closely aligned with the observed peak than the attention-derived peak.

\section{Discussion}

The central empirical finding is a dissociation between forecast-output timing
and horizon-level attention. The forecast peak is closely aligned with the
observed peak, whereas the attention-derived horizon reference is not. The
peak-neighborhood mass analysis further shows that this discrepancy is not
merely an artifact of taking the attention argmax: in the primary run,
attention is only weakly concentrated around the observed or predicted peak
and is similarly concentrated around a weekly-naive peak. Thus, the specific
interpretation of the attention profile as indicating where the forecast peak
is formed is not supported in this setting.

At the same time, $\psio$ is not unstructured noise. Its profile is highly
consistent across test windows, indicating a stable temporal structure rather
than arbitrary variation. This distinction is important: failure to explain
demand-peak timing does not imply that the attention mechanism itself is
meaningless or unused by the model.

The architecture provides an important clue for interpreting this result.
The analyzed attention distribution is generated from a single CLS query
summarizing the historical demand state, which attends jointly to 24 future
exogenous horizon tokens. It therefore represents a global horizon reference
rather than a horizon-specific attribution of individual forecast outputs.
Accordingly, $\psio$ need not behave as a selector of the future demand-peak
horizon. It may instead reflect how the historical state references future
exogenous conditions while constructing the internal representation used for
forecasting. Consistent with this interpretation, removing variation in the
future weather features (by replacing them with training-set means)
substantially flattens the attention profile. The
structured pattern is therefore sensitive to future weather variation and
cannot be explained by fixed horizon position alone.

Structurally, there is no one-to-one correspondence between the attention
weight at horizon $h$ and the forecast output $\hat{y}(h)$: a single CLS query
attends jointly to exogenous horizon tokens, and the resulting representation
is used to generate the full 24-step forecast. Moreover, the attended tokens
contain weather and calendar information rather than future demand itself. A
horizon that is informative for forecast formation therefore need not coincide
with the demand-peak horizon.

This alternative interpretation also changes how the forecast--attention
mismatch should be understood. The lack of alignment between the attention
peak and the demand peak is not, by itself, evidence that the attention
mechanism is uninformative. Future exogenous conditions may also affect demand
with temporal delays, so exact horizon-to-peak correspondence should not be
expected in general. Our conclusion is therefore limited to the direct,
peak-selective interpretation examined here: the observed attention profile
does not provide evidence that the attended horizon identifies where the
forecast peak is formed. More generally, an internally structured attention
pattern can be misinterpreted when it is mapped directly onto an output event
that the mechanism was not necessarily designed to select. The present sensitivity
analysis is consistent with an internal horizon-reference role for integrating
future exogenous information.

Two further implications follow. First, the comparable peak-timing summaries
of the forecast and the weekly seasonal-naive predictor are consistent with
strong recurring temporal structure in this task. This makes it
unnecessary to expect the exogenous cross-attention profile itself to be
peak-selective: future weather and calendar information may contribute to
forecast formation without its attention maximum coinciding with the
demand-peak horizon. Determining how peak-timing information is distributed
across the historical and exogenous pathways would require a separate
intervention. Second, from a practical standpoint, the attention profile over
future horizons should not be presented as a rationale for where the predicted
peak lies unless that interpretation has been validated against the peak event
itself. Output-aligned attribution methods could provide one possible direction
for such validation.

From an evaluation-validity perspective, the key question is therefore not
simply whether attention exhibits structure or responds to its inputs, but
whether the particular interpretation assigned to that structure is supported
by observable model behavior. Stability, input sensitivity, reproducibility,
and visual coherence are insufficient on their own. An attention pattern can
exhibit these properties while remaining weakly aligned with the output event
it is claimed to explain. This distinction matters in particular when
attention maps are read visually, because a stable and coherent pattern can
look explanatory even when the output-level interpretation attached to it is
unsupported. The secondary failure-diagnostic analysis further
shows that forecast--attention disagreement is not, in this setting, a useful
indicator of peak-timing failure, reinforcing that diagnostic utility and
explanatory validity are distinct questions
(Appendix~\ref{app:failure_diagnostic}).

The random-seed analysis strengthens the empirical part of this conclusion.
Although the precise attention peak error and local peak concentration vary
with initialization, the direction of the central result does not: in all
three runs, forecast peak timing is more closely aligned with the observed
peak than the attention-derived reference. The dissociation is therefore not
specific to a single initialization.

\section{Limitations}

This study remains a single-dataset and single-imaging-method (GAF) case study.
The reported values should therefore be interpreted as evidence from this
setting rather than as a universal statement about attention-based forecasting
models.
Our claim is intentionally limited in scope: the present experiments show
that such a dissociation can occur in a realistic forecasting setting, but do
not establish how prevalent it is across forecasting tasks. Although the
central dissociation is reproduced across three random seeds, broader
evaluation across datasets, forecasting architectures, and alternative
time-series image representations is therefore required to assess generality. 

A further open question concerns how geometric re-representation of
temporal structure influences the internal representations learned by
forecasting models. Although this study uses a GAF-based historical
representation, it does not examine which temporal relationships are
preserved, exposed, or obscured by this transformation, or how they are
reflected in the learned internal state. Future work should investigate
these relationships to better understand how temporal structure is
represented and used during forecasting. Consistent with the evaluation
principle emphasized in this study, interpretations of such internal
structures should be validated against observable forecast behavior
rather than inferred from structural properties alone.

Peak timing is defined by an argmax, which can be unstable for diffuse
profiles; the peak-neighborhood mass analysis mitigates but does not
eliminate this ambiguity. The weather replacement experiment is a
sensitivity probe rather than a causal intervention. It shows that the
attention structure responds to changes in future weather inputs and
is consistent with a horizon-reference interpretation of $\psio$.
However, mean replacement may also shift the input distribution, and
alternative internal mechanisms could produce a similar pattern.
Establishing the functional role of the attention mechanism would
require stronger intervention or representation-level analyses.
Because future weather is taken from reanalysis records rather than
operational forecasts, whether the same behavior persists under
weather-forecast uncertainty remains untested.

The present findings are also specific to an architecture in which a single
CLS query attends jointly to future exogenous horizons. Because this attention
operates over exogenous inputs rather than directly over forecast outputs,
different attention--output relationships may arise in architectures with
horizon-specific queries, decoder-style prediction tokens, or other
output-aligned attention mechanisms.

The failure-diagnostic analysis reported in
Appendix~\ref{app:failure_diagnostic} is limited to the present test setting
and to peak-time error as the forecast-failure criterion. The absence of a
meaningful association between forecast--attention disagreement and peak-time
error should therefore not be interpreted as ruling out diagnostic utility
for other error measures, datasets, or forecasting settings. In addition,
adjacent forecast windows share portions of their 168-step history. The windows
should therefore not be interpreted as fully independent samples, and the
Wilcoxon tests are treated as supportive paired comparisons rather than the
sole basis of the conclusion. Finally, the observed-peak attention lift varies
across random seeds ($1.12$--$1.56$). The robust claim is therefore the
forecast--attention dissociation itself, not that the precise degree of
peak-neighborhood concentration or the detailed attention profile is invariant
to initialization.

\section{Conclusion}

We examined whether horizon-level cross-attention provides a valid explanation
of predicted demand-peak timing in an imaging-ViT load forecaster. The forecast
and attention show a clear temporal dissociation: the forecast peak is closely
aligned with the observed peak, while the attention-derived peak is
substantially less aligned. This
mismatch persists when attention is evaluated through local peak-neighborhood
mass rather than argmax alone and is reproduced across three independently
trained runs initialized with different random seeds.

At the same time, the attention profile is highly consistent across test
windows and becomes nearly uniform when variation in future weather features is
removed. The attention mechanism is therefore neither unstructured noise nor
simply a fixed positional template. Together with the model architecture, this
behavior is consistent with $\psio$ serving as an internal horizon reference
for future exogenous information rather than as a selector of the demand-peak
horizon.

Thus, the main lesson is not that attention is intrinsically uninformative,
but that the functional role of an attention mechanism and the output-level
explanation attributed to it are distinct questions. Attention structure,
input sensitivity, diagnostic utility, and output-level explanatory validity
should therefore be evaluated as distinct properties. A structured,
reproducible, and input-sensitive attention pattern may be internally
meaningful while still failing to support the behavioral interpretation
suggested by its visualization. Attention-based explanations should therefore
be validated against the specific output behavior they are claimed to explain
rather than inferred from attention structure alone.

\section*{Acknowledgments}

The authors thank Asuka Takeuchi for helpful discussions and insights.

\clearpage
\appendix

\section{Representative Demand Windows}
\label{app:representative_waveforms}

\begin{figure}[ht]
  \centering
  \includegraphics[width=0.90\linewidth]{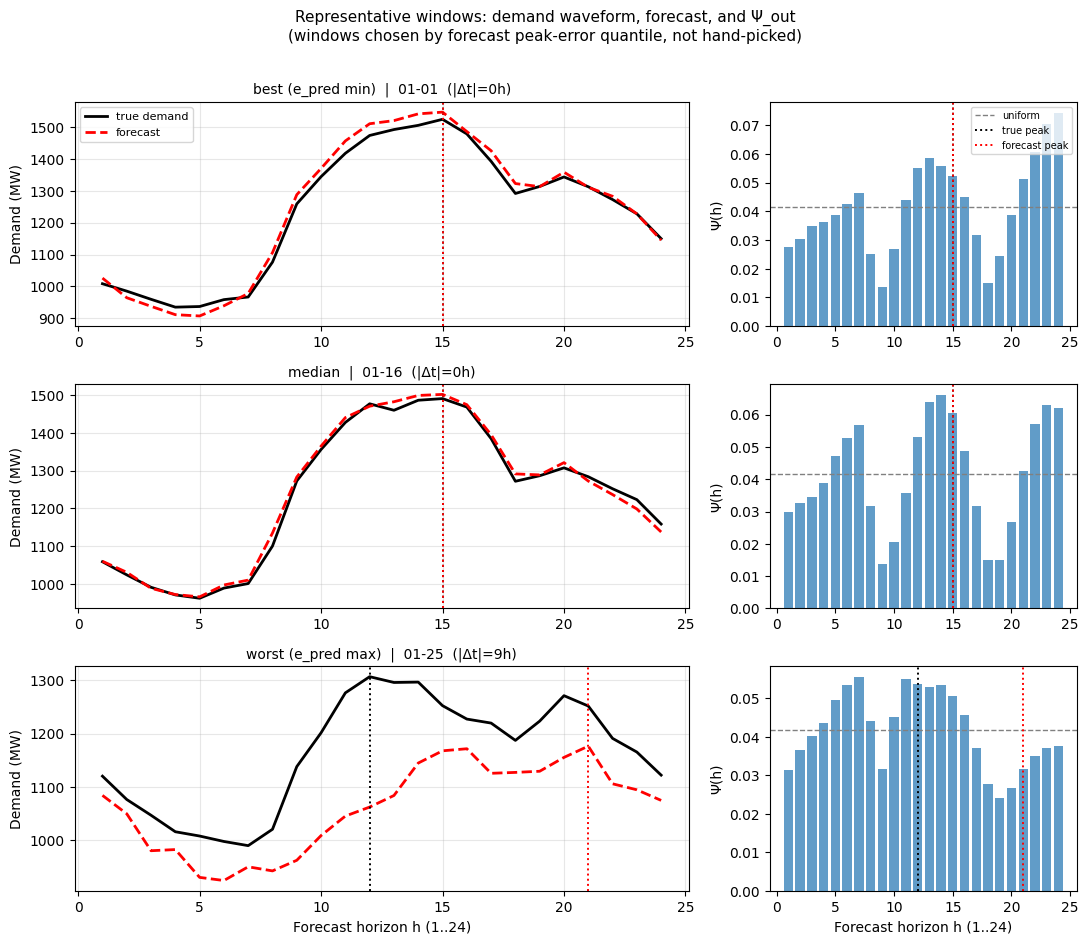}
  \caption{
  Representative day-aligned windows selected by forecast peak-time error:
  best, median, and worst. The left column shows the observed demand and
  forecast, with vertical lines marking their peak horizons. The right column
  shows the corresponding horizon-level attention profile $\psio(h)$ and the
  uniform reference level. The examples illustrate the dissociation between
  forecast peak timing and the attention-derived horizon reference.
  }
  \label{fig:representative_waveforms}
\end{figure}

Figure~\ref{fig:representative_waveforms} shows representative day-aligned
windows selected mechanically by forecast peak-time error quantiles. We show the
best, median, and worst windows according to forecast peak-time error. These
examples provide qualitative context for the main analysis and illustrate that
forecast peak timing can remain accurate even when the horizon-level attention
profile is diffuse and not peak-selective.

\section{Forecast--Attention Disagreement as a Failure Diagnostic}
\label{app:failure_diagnostic}

As a secondary analysis, we asked whether disagreement between the forecast
peak and the attention-derived peak could serve as a diagnostic signal for
unreliable peak forecasts. For each test window, we define the disagreement as
\begin{equation}
  d_{\Psi,\mathrm{pred}}
  =
  |\hpsi-\hpred|.
\end{equation}
We then measure its association with forecast peak-time error
$e_{\mathrm{pred}}$ across the test windows.

The association is weak and not statistically meaningful
(Spearman $r=0.08$, $p=0.65$). Thus, in this test setting, a large disagreement
between the attention-derived horizon and the forecast peak does not provide a
useful indicator of peak-timing failure. This result is secondary to the main
analysis: it does not imply that attention-based quantities cannot provide
diagnostic information under other definitions of forecast error or in other
forecasting settings.

\section{Implementation and Reproducibility Details}
\label{app:implementation}

The forecasting model uses a ViT-B/16 backbone with a 168-step
history window and a 24-step forecasting horizon. The input time
series is transformed into a Gramian Angular Field (GAF) image and
resized to $224\times224$ pixels. The ViT hidden dimension is 768.
Future exogenous variables are represented as horizon-wise tokens
and fused with the image-derived CLS state through an 8-head
cross-attention block inserted after transformer block 8.
The updated CLS token is recombined with the image patch tokens
and passed through the remaining ViT blocks before the final
CLS representation is used for 24-step regression.
The cross-attention block and regression head use dropout 0.1.

Models are trained using AdamW with mean squared error as the
training objective. The learning rate is $1\times10^{-5}$.
Training uses a batch size of 32, a maximum of 50 epochs, and
early stopping with patience 10 based on validation RMSE.
The ViT backbone is frozen during the first epoch and then jointly
fine-tuned. For the ViT-based model, weight decay is 0.05.

Training data extend from the beginning of the dataset through
November 19, 2018, with validation performed from November 20
through December 31, 2018. January 2019 is held out for testing.
All normalization statistics for the target and exogenous variables
are estimated exclusively from the training period and then kept
fixed for validation and test inference. Random seeds are explicitly
set for Python, NumPy, and PyTorch, with deterministic CUDA behavior
enabled.

\subsection{Seed-robustness analysis}
\label{app:seed_protocol}

The random-seed robustness analysis is performed from the saved
test predictions and horizon-level attention profiles for seeds
0, 1, and 2. We restrict the analysis to day-aligned origins at
23:00 during January 2019. Before comparing seeds, we verify
programmatically that the complete set of evaluation timestamps is
identical across runs. The common evaluation set contains 30 origins.

For each seed, forecast-peak and attention-peak errors are recomputed
from the saved outputs. A two-sided paired Wilcoxon signed-rank test
is then performed separately for each seed. We additionally compute
the observed-peak neighborhood lift and cross-window attention
consistency.

\subsection{Existing assets}
\label{app:assets}

The Panama electricity-demand dataset used in this study is publicly
available through Mendeley Data (Version 1,
DOI: 10.17632/byx7sztj59.1) under the CC BY 4.0 license.
The ViT-B/16 backbone uses the
\texttt{vit\_base\_patch16\_224} implementation and pretrained weights
provided through the \texttt{timm} library. The pretrained configuration
used is \texttt{augreg2\_in21k\_ft\_in1k}, distributed under the
Apache 2.0 license.

\end{document}